\documentclass{article}
\usepackage{spconf,amsmath,epsfig}

\usepackage[inline]{enumitem}
\usepackage{cite}
\usepackage{amsmath,amssymb,amsfonts}
\usepackage{algorithmic}
\usepackage{graphicx}
\usepackage{textcomp}
\usepackage[table, dvipsnames]{xcolor}
\usepackage{subcaption}
\usepackage{array}
\usepackage[most]{tcolorbox}
\usepackage{sectsty}
\usepackage{float}

\newcommand{\figref}[1]{Figure~\ref{#1}}

\newcommand{\secref}[1]{Section~\ref{#1}}

\newcommand{\tabref}[1]{Table~\ref{#1}}
\newcommand{\code}[1]{\texttt{#1}}

\title{Convolution Based Spectral Partitioning Architecture for Hyperspectral Image Classification}

\begin{document}
%
\maketitle
%
%
%
\begin{abstract}
Hyperspectral images (HSIs) can distinguish materials with high number of spectral bands, which is widely adopted in remote sensing applications and benefits in high accuracy land cover classifications. However, HSIs processing are tangled with the problem of high dimensionality and limited amount of labelled data. To address these challenges, this paper proposes a deep learning architecture using three dimensional convolutional neural networks with spectral partitioning to perform effective feature extraction. We conduct experiments using Indian Pines and Salinas scenes acquired by NASA Airborne Visible/Infra-Red Imaging Spectrometer. In comparison to prior results, our architecture shows competitive performance for classification results over current methods.\stepcounter{footnote}\footnotetext{Work performed at Imperial College London.}

\end{abstract}

\section{Introduction}
Hyperspectral images (HSIs) contain spectrum information for each pixel in the image of a scene, where each spatial pixel is a spectral vector composed of hundreds of contiguous narrow electromagnetic bands reflected or radiated by the detecting materials. HSI classification involves assigning a categorical class label to each unlabelled pixel based on the corresponding spectral and/or spatial feature~\cite{b1}. With the advent of new hyperspectral remote sensing instruments and their increased temporal resolutions, the amount of high dimensional hyperspectral data is increasing. This results in new practical and theoretical problems due to the high dimensionality where traditional algorithms developed for multi-spectral imagery may no longer be suitable.

Convolutional neural networks (ConvNets/CNN) have shown potential in hyperspectral imagery classification as they use extensive numbers of parameters for feature learning~\cite{b6}. In spite of the potentiality, high variability spectral signature properties of HSIs complicates the corresponding CNN designs; Time consuming and expensive manual labelling of HSIs has limited the number of training samples. These problems have obstructed and reduced the predictive power of CNN models.


Existing CNN models for HSI classification are often based on one dimensional or two dimensional CNN architectures. The former set of models are adapted for spectral feature learning~\cite{b5}, while the later sets of models explore local spatial feature learning at each band~\cite{chen16a}. These models show deficiency in performing feature extraction on multiple dimensions. 
Three dimensional CNN is exploited by researchers~\cite{3DCNN_IGARSS} and yet the network structure is often too sophisticated for deployment on real-time embedded devices such as CPU, FPGA or even GPU.
Traditional machine learning methods including Logistic Regression~\cite{b4} and Kernel-based SVM~\cite{b3} are proposed by researchers but it has been reported that classification accuracies is inferior and less favourable.


In this paper, we propose a novel 3D CNN architecture that tackles these challenges by 
\begin {enumerate*} [label=\itshape\alph*\upshape)]
\item \textit{spectral partitioning within the network to process pixels in each dimension,} and by 
\item \textit{efficient spatial-spectral feature extraction } 
\end {enumerate*} 
Essentially, the proposed model first performs a spatial transformation via 2D convolution. The transformed image is partitioned on the spectral level and split into segments for efficient processing. 3D convolution is then applied to each segment. Finally, convoluted segments are concatenated and summarized with fully-connected layers with dropout as regularization to prevent over-fitting. 
Our architecture is shown to be robust and stable where the model is more accurate compared to BASS~Net~\cite{b5} and 2D-CNN~\cite{chen16a}. We also demonstrate and train our architecture using as minimal training samples as possible, which in our case, 20\% and 10\% of labelled data from Indian Pines and Salinas dataset respectively. All source code can be found in section III.

\section{Proposed Methodology}

\begin{figure*}
\centerline{\includegraphics[scale=0.4]{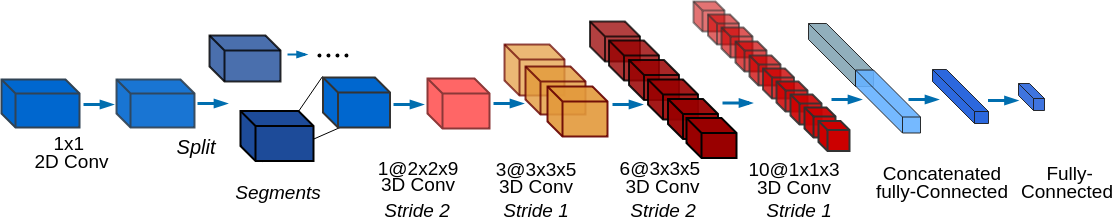}}
\vspace{-0.225cm}
\caption{An illustration of the proposed neural network architecture for HSI classification. The diagram is showing the operation of one of the two segments only. Both segments follow identical 3D convolution operation.}
\label{fig:network}
\end{figure*}

Hyperspectral images are typically represented as a data cube in dimension $(x, y, \lambda)$, where $x$ and $y$ represent spatial dimensions with space information of pixels, and $\lambda$ represents the third dimension with a spectral vector that can be used for distinguishing different materials and objects. To improve the classification accuracy and reduce the number of training samples, we propose the use of 3D CNN to perform spatial-spectral feature extraction with spectral partitioning over $(x, y, \lambda)$ dimensions.

\subsection{Three Dimensional Convolutional Neural Networks}
As mentioned, 3D CNN is capable of capturing features from both spatial and spectral dimensions when compared to traditional 1D and 2D CNNs. The value of a neuron at a given position $(x,y,z)$ is denoted as:
\begin{equation}
\small
v^{xyz}_{ij}=\sigma\bigg(\sum_{m}\sum_{h=1}^{H_i-1}\sum_{w=1}^{W_i-1}\sum_{r=1}^{R_i-1}k^{hwr}_{ijm}v^{(x+h)(y+w)(z+r)}_{(i-1)m}+b_{ij} \bigg)
\label{eq:conv3d}
\end{equation}
where $i$ indicates the current layer, $m$ indexes the feature map in the $(i-1)$th layer connected to the current feature map. $H_i$ and $W_i$ represent the height and width of the kernel, $R_i$ represents the depth of the kernel towards the spectral dimension, $k^{hwr}_{ijm}$ is the value of $(x,y,z)$ on feature $m$, $b_{ij}$ is the bias term of feature $j$ on layer $i$ and $\sigma$ indicates the activation function.

\subsection{Proposed Architecture}
\label{sssec:num1}

\begin{figure}[htbp]
\centering
\includegraphics[scale=0.3]{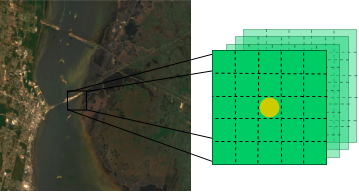}
\vspace{-0.2cm}
\caption{Cubic data extract from a HSI as input to proposed CNN architecture.}
\label{fig:HSIneighbours}
\end{figure}

Our architecture takes an input of a small image cube from  hyperspectral data. The small cube is taken from 24 neighbours of its central pixel with patch size $=5$, where the central pixel is required for classification (Fig.~\ref{fig:HSIneighbours}). In other words, the input image cube has size $5\times 5 \times N_b$, where $N_b$ is the spectral size of the hyperspectral image. From our prior experiment, a patch size $=5$ contributes to the finest result in classification when compared to patch size $=3$ or $7$.

This cubic data is then fed to the proposed CNN architecture as briefly illustrated in \figref{fig:network}. The architecture begins with one 2D convolution layer using a $1 \times 1$ filters for linear transformation of each input spectral band. 
\textit{Spectral partitioning} is then performed on the convoluted image, where the image is split into two segments with a non-overlapping spectral axis. Each segments are followed by a stack of four 3D convolutions sequentially with different kernel size. Operations of each segments are identical, kernel weights and biases of the stack of convolutions are shared across the segments to reduce total number of parameters.


In the first 3D CNN layer, it has one convolution filter of size $2\times2\times9$ with sub-sampling stride $2$ on $\lambda$. Consider the data cube contains redundant information, the purpose of this layer is to learn the low-level features and reduce the high-dimensionality. The second layer has three filters with size $3\times3\times5$ and stride $1$. This layer is designed for learning local mid-level features of  HSI. The third layer has five $3\times3\times5$ kernels with stride $2$ to further reduce the dimensionality on the spectral level. The final 3D-CNN layer consists of 10 $1\times1\times3$ kernels with stride $1$, which aims at extracting high-level spectral features.

The convoluted segments are concatenated and fed to two Fully-Connected layers (FC); The first one has 120 neuron units and the second one has number of units equal to the number of categorical classes of the dataset.
Note that Rectify Linear Units (ReLu) are employed as activation functions on every layer except the final layer, where Softmax function is applied for classification. The final label for the pixel is given by the $\arg \max$ from the Softmax function.



\subsection{Optimisation, Learning, Training and Inference}

With limited dataset provided in the training stage, over-fitting can become a serious problem where classification performance is good on the training set, but low on the testing set. Therefore, the training is regularised by applying dropout at fully-connected layer, with ratio set to $p=0.5$.

We carry out the training by minimizing the Cross-Entropy object function using Adaptive Moment Optimization gradient descent (Generally known as Adam). The adam is set to have initial learning rate $5\times10^{-4}$. The batch size is set to have 50 for each training iteration and require 650 epochs for convergence.  

The segments described in \secref{sssec:num1} are placed sequentially during the training stages; During the inference stage, segments can be adjusted to align in sequential, parallel or pipeline, and can be customized with regards to runtime constraints. In this paper, we demonstrate our model by placing segments in parallel.





\section{Experiments \& Results}

\subsection{Dataset and Preprocessing}
The Indian Pines Scene and Salinas Scene datasets, which were acquired by Airborne Visible/Infrared Imaging Spectrometer (AVIRIS) over Northwestern Indiana and Salinas-Valley, California, are used in this experiment. Indian Pines scene provides 224 spectral channels in the wavelength ranges from 0.4 to 2.5$\mu$m. 200 channels remain after discarding the water absorption regions. We remove some classes in this dataset due to insufficient samples. Similar to Indians Pines scene, Salinas scene also consists of 224 spectral bands. 20 channels are discarded for correction and only 204 channels remain for the experiment. 

We randomly select 20$\%$, 5$\%$ and 75$\%$ from each class as our training data, validation data and testing data respectively for Indian Pines dataset. Similarly, we randomly select 10$\%$, 5$\%$ and 85$\%$ as the training, validation and testing set for Salinas dataset. All pixel values are calibrated through normalization via the following transformation $z_i = \frac{x_i-\min(x)}{\max(x)-\min(x)}$.
where $x=(x_1...x_n)$ represents all pixel values of an HSI. 

\subsection{Classification Result}
We implement our model using TensorFlow, and Scikit-learn and it is trained on a single NVIDIA GeForce GTX 1080Ti GPU. Source code can be viewed on Github as an open source project~\footnote{\color{Maroon} https://github.com/custom-computing-ic/SpecPatConv3D-Network}.

\begin{table}
    \centering
    \small
     \caption{Comparison of proposed network versus other methods in classification accuracy (\%) on Indian Pines dataset.}
    \begin{tabular}{|l|c|c|c|}
    \hline
\textbf{\textit{Class}} & \textbf{\textit{BASS Net}}& \textbf{\textit{Conv2D}}& \textbf{\textit{Our Model}} \\ \hline
         
        Corn-notil  & 94.77 & 91.97 & 98.23  \\  \hline
        Corn-mintill & 95.02 & 95.98 & 97.58\\  \hline
        Corn & 94.94 &  98.88 & 96.07  \\  \hline
        Grass-pasture  & 99.17 & 99.72 & 99.17 \\  \hline
        Grass-trees  & 99.63 & 99.08 & 99.27 \\  \hline
        Hay-windrowed  & 100.00 & 100.00 & 100.00 \\  \hline
        Soybean-notil  & 96.29 & 98.77 & 98.77 \\  \hline
        Soybean-mintil  & 97.88 & 98.26 & 96.36 \\  \hline
        Soybean-clean  & 97.52 & 90.54 & 98.20 \\  \hline
        Woods  & 97.68 & 98.63 & 99.05 \\  \hline
        BGTD  & 98.28 & 94.48 & 97.93 \\  \hline \hline
 \textbf{\code{OA}} & 97.38$\pm$0.2 & 97.00$\pm$0.1 & \cellcolor{red!25} 97.98$\pm$0.2  \\
\hline
 \textbf{\code{AA}} & 97.23$\pm$0.2 & 96.76$\pm$0.1 & \cellcolor{red!25} 98.92$\pm$0.2 \\
 \hline
    \end{tabular}
    \label{tab1:ip}
   
\end{table}
\begin{table}
\small
\caption{Comparison of proposed network versus other methods in classification accuracy (\%) on Salinas dataset.} 
\vspace{-0.35cm}
\begin{center}
\begin{tabular}{|p{2.68cm}|c|c|c|}
\hline
\textbf{\textit{Class}} & \textbf{\textit{BASS Net}}& \textbf{\textit{Conv2D}}& \textbf{\textit{Our Model}} \\ \hline
Brocoli green \newline weeds 1 & 100.00 & 99.78 & 100.00  \\  \hline
Brocoli green\ newline weeds 2 & 99.72 & 100.00 & 99.82  \\  \hline
Fallow   & 100.00 & 100.00 & 100.00  \\  \hline
Fallow rough plow & 99.84 & 100.00 & 100.00  \\  \hline
Fallow smooth   & 99.83 & 98.91 & 99.50  \\  \hline
Stubble   & 100.00 & 100.00 & 100.00  \\  \hline
Celery   & 99.69 & 99.56 & 99.91  \\  \hline
Grapes untrained & 97.02 & 91.37 & 94.20  \\  \hline
Soil vinyard develop & 100.00 & 99.9 & 99.60  \\  \hline
Corn senesced \newline green weeds & 96.78 & 98.01 & 97.46  \\  \hline
Lettuce romaine 4wk  & 95.89 & 97.37 & 97.47  \\  \hline
Lettuce romaine 5wk  & 98.34 & 100.00 & 99.59  \\  \hline
Lettuce romaine 6wk  & 100.00 & 100.00 & 100.00 \\  \hline
Lettuce romaine 7wk   & 99.26 & 99.47 & 99.37  \\  \hline
Vinyard untrained & 84.31 & 87.96 & 96.52  \\  \hline
Vinyard vertical \newline trellis  & 99.13 & 99.07 & 99.32  \\  \hline
\hline
 \textbf{\code{OA}} & 96.84$\pm$0.1 & 96.27$\pm$0.1 &  \cellcolor{red!25} 98.73$\pm$0.1  \\
\hline
 \textbf{\code{AA}} & 98.12$\pm$0.1 & 98.21$\pm$0.1 & \cellcolor{red!25}98.92$\pm$0.1 \\
\hline
\end{tabular}
\label{tab1:salinas}
\end{center}
\end{table}

We compare the performance of our approach to two other recent techniques: BASS~Net~\cite{b5} and 2D-CNN~\cite{chen16a} to obtain the overall accuracy (\code{OA}) and the average per-class accuracy (\code{AA}) using the same dataset. Accuracy values are calculated based on the average of five classification runs.
\tabref{tab1:ip} and \tabref{tab1:salinas} summarise the results on \code{OA} and \code{AA} for different approaches  on Indian Pines and Salinas datasets.

\begin{figure*}[t!]
\centering

\begin{center}
\includegraphics[width=0.15\textwidth]{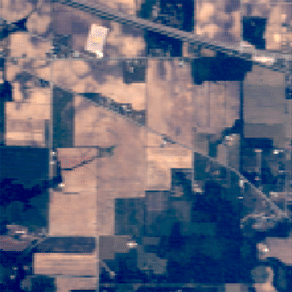} 
\includegraphics[width=0.15\textwidth]{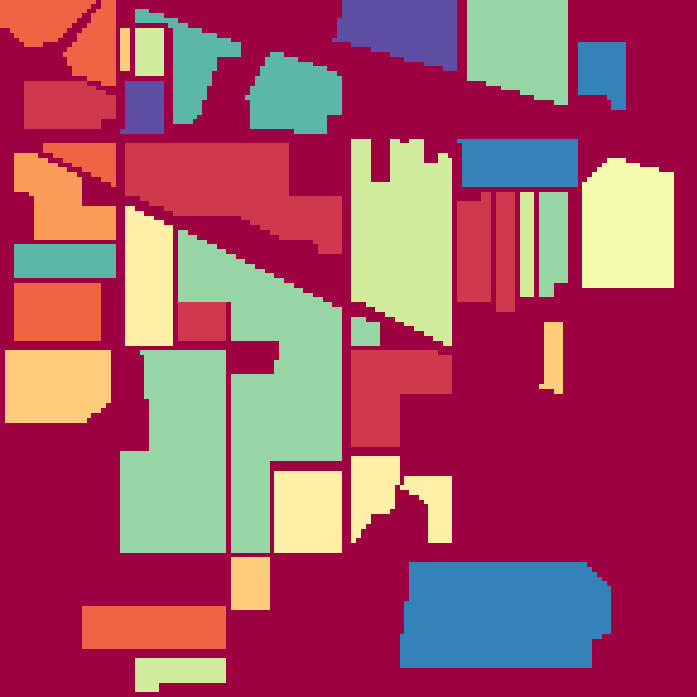}
\includegraphics[width=0.15\textwidth]{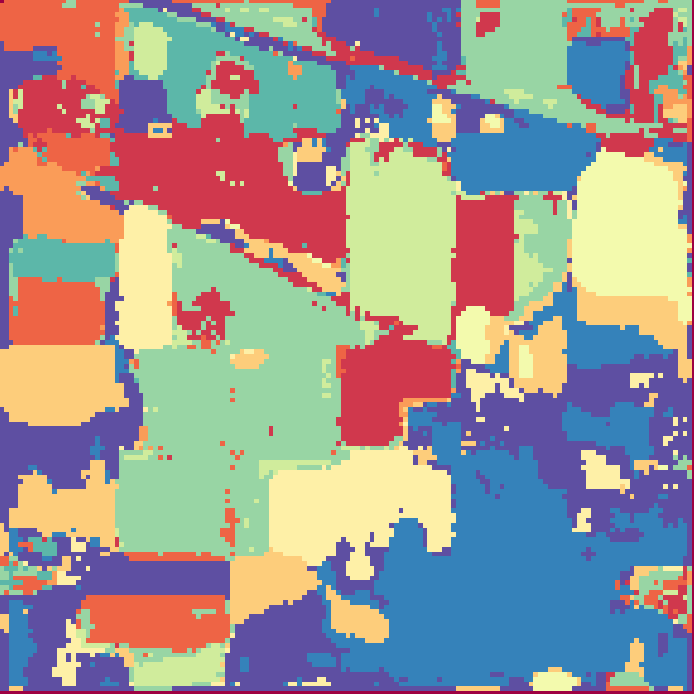} 
\includegraphics[width=0.15\textwidth]{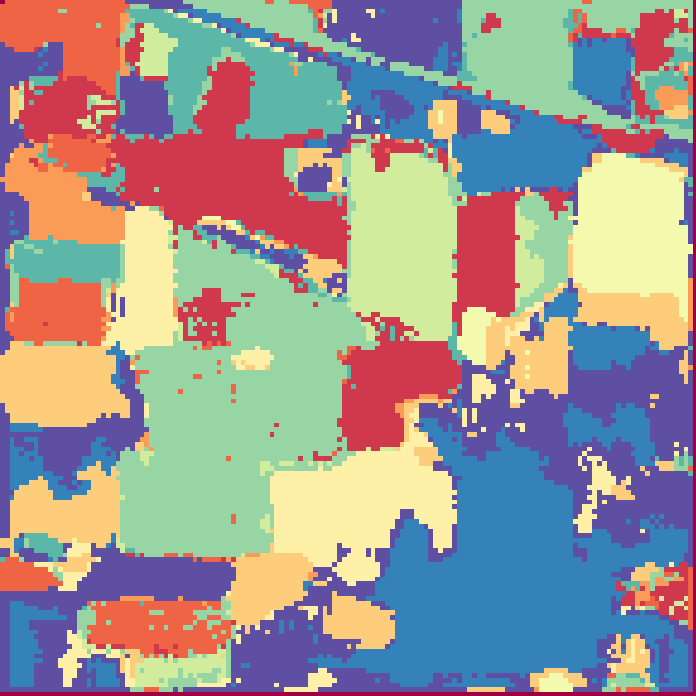} 
\includegraphics[width=0.15\textwidth]{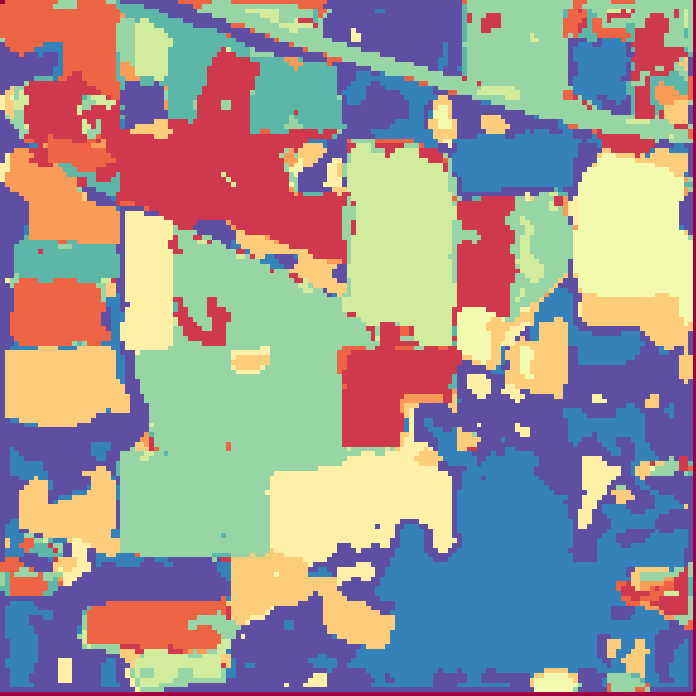} 
\includegraphics[height=0.15\textwidth]{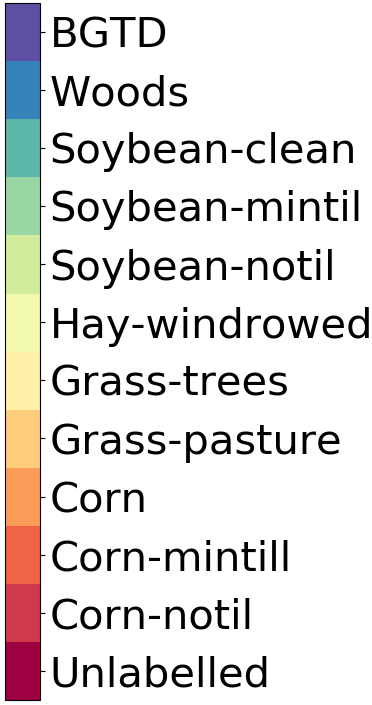} 
\\
\end{center} \vspace{-0.5cm}
\caption{ Indian Pines dataset: True colour composite, Ground truth image, Classification map of BASS-Net, Conv2D and Our method. (Left to right)}
\label{fig:result1}

\vspace{-0.275cm}
\begin{center}
\includegraphics[height=0.22\textheight]{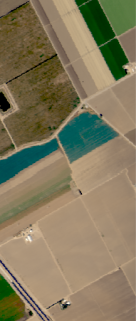} 
\includegraphics[height=0.22\textheight]{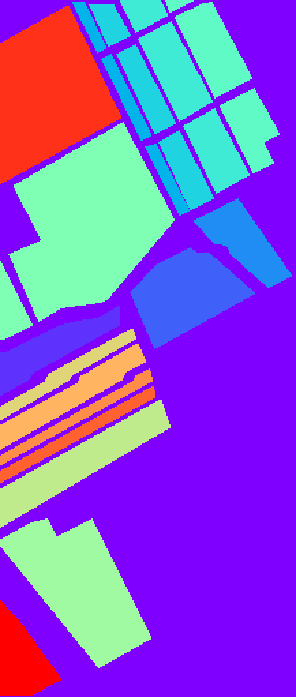} 
\includegraphics[height=0.22\textheight]{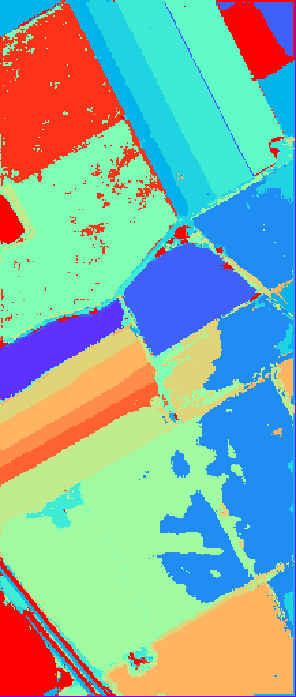} 
\includegraphics[height=0.22\textheight]{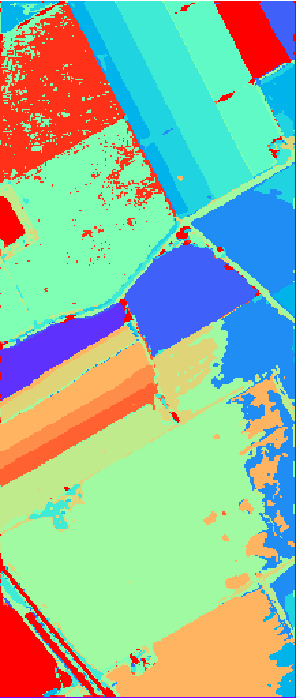} 
\includegraphics[height=0.22\textheight]{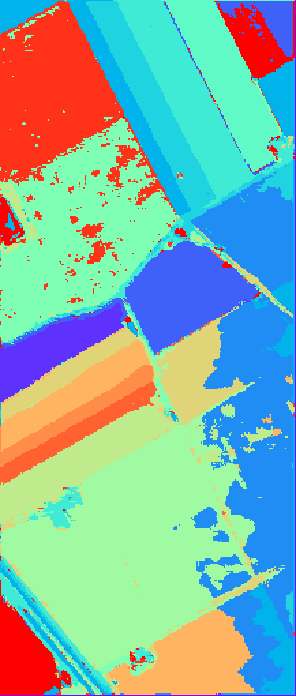} 
\includegraphics[height=0.22\textheight]{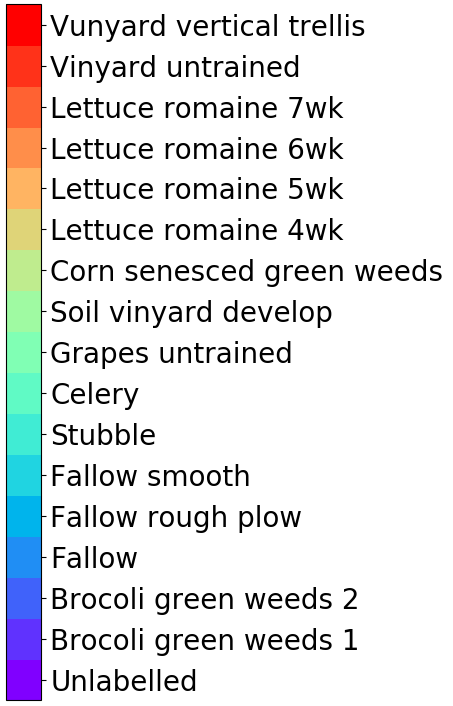} 
\\
\end{center} \vspace{-0.5cm}
\caption{Salinas dataset: True colour image, Ground truth image, Classification map of BASS-Net, Conv2D and Our method. (Left to right)}
\label{fig:result2}

\end{figure*}

Our model obtains the best results compared to other models, with 98.0$\%$ and 98.7$\%$ in overall accuracy on Indian Pines scene and Salinas scene respectively. This demonstrates the superiority of the proposed 3D CNN for hyperspectral classification, where the corresponding classification maps of the proposed model clearly distinguish the boundaries between different classes (\figref{fig:result1} and \figref{fig:result2}).

\begin{figure}[h]
\begin{subfigure}{.24\textwidth}
  \centering
  \includegraphics[height=1.7in]{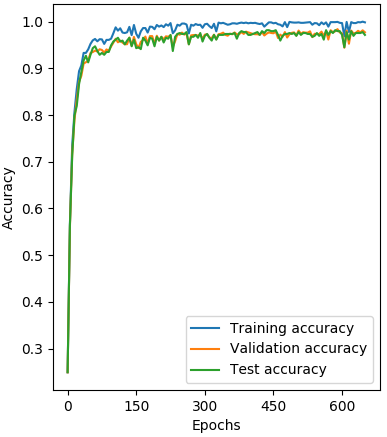}
  \vspace{-0.25cm}
  \caption{}
\end{subfigure}%
\begin{subfigure}{.24\textwidth}
  \centering
  \includegraphics[height=1.7in]{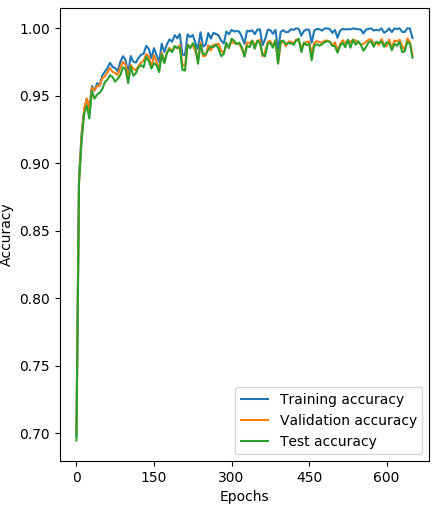}
  \vspace{-0.25cm}
  \caption{}
\end{subfigure}\vspace{-0.35cm}
\caption{The changes in accuracy against the epochs for (a) Indian Pines dataset, (b) Salinas dataset.}
\label{fig:acc}
\vspace{-0.3cm}
\end{figure}

Finally, the changes in accuracy against the epochs for the training, validation and testing data is shown in \figref{fig:acc} for both datasets. Intuitively, the network converges after 100 epochs and reaches to an optimal point with a stable rate after 600 epochs for the training data. After the convergence, the testing and validation accuracy are closed to training accuracy. This demonstrates the robustness of our model where the over-fitting problem is diminished and the features on HSI are extracted and learned by the proposed model.

\section{Conclusion}
This paper proposes an architecture for HSI classification, with spectral partitioning to reduce dimensionality and spatial-spectral features extracted by 3D CNN. Experiments show that our method outperforms comparable methods regarding classification accuracy while using a fewer amount of training data.

Currently, the proposed architecture consists of multiple layers of 3D CNN, which is still very computationally intensive. Future work will include complexity reduction of the model. For example, extending the network with other types of layers such as depth-wise convolution and migrating the design onto embedded devices such as FPGA to perform real-time processing.

{\small\section{Acknowledgememt}
The authors are grateful for the support by Intel, United Kingdom EPSRC (grant numbers EP/I012036/1, EP/L00058X/1, EP/L016796/1, EP/N031768/1), European Union Horizon 2020 Research and the Lee Family Scholarship.
}
\bibliographystyle{IEEEbib}
{\small\bibliography{reference}}

\end{document}